\documentclass[letterpaper, 10 pt, conference]{ieeeconf} 
\IEEEoverridecommandlockouts                              
\usepackage{graphicx}
\usepackage{amsmath}
\usepackage{amssymb}
\usepackage[table,xcdraw,dvipsnames]{xcolor}
\usepackage{booktabs}
\usepackage{url}           
\usepackage{amsfonts}      
\usepackage{nicefrac}      
\usepackage{pifont} 
\let\labelindent\relax
\usepackage{enumitem}
\usepackage{caption}
\usepackage{url}
\usepackage{cuted} 
\usepackage[colorlinks,citecolor=ForestGreen]{hyperref}

\newcommand{\myParagraph}[1]{\noindent \textbf{#1} ---}

\definecolor{ColNeural}{HTML}{EBB48A}
\definecolor{ColClassical}{HTML}{A5C2E3}
\definecolor{ColHL}{HTML}{CADFB8}

\newcommand{\xmark}{\ding{55}}%

\definecolor{Gray}{gray}{0.85}
\definecolor{GrayBorder}{gray}{0.65}
\newcolumntype{a}{>{\columncolor{GrayBorder}}l}
\newcolumntype{b}{>{\columncolor{Gray}}c}
\newcolumntype{d}{>{\columncolor{GrayBorder}}c}

\title{\LARGE \bf
Learning whom to trust in navigation: dynamically switching between classical and neural planning}

\author{Sombit Dey$^{1}$, Assem Sadek$^{2}$, Gianluca Monaci$^{2}$, Boris Chidlovskii$^{2}$ and Christian Wolf$^{2}$
\thanks{$^{1}$Sombit Dey is with ETH Z\"{u}rich. Work done while at Naver Labs Europe.
        {\tt\small somdey@ethz.ch}}%
\thanks{$^{2}$ All other authors are with Naver Labs Europe, Meylan, France. {\tt\small firstname.lastname@naverlabs.com}}%
}

\begin{document}

\maketitle


\thispagestyle{empty}
\pagestyle{empty}

\begin{abstract}
Navigation of terrestrial robots is typically addressed either with localization and mapping (SLAM) followed by classical planning on the dynamically created maps, or by machine learning (ML), often through end-to-end training with reinforcement learning (RL) or imitation learning (IL). Recently, modular designs have achieved promising results, and hybrid algorithms that combine ML with classical planning have been proposed. Existing methods implement these combinations with hand-crafted functions, which cannot fully exploit the complementary nature of the policies and the complex regularities between scene structure and planning performance.

Our work builds on the hypothesis that the strengths and weaknesses of neural planners and classical planners follow some regularities, which can be learned from training data, in particular from interactions. This is grounded on the assumption that, both, trained planners and the mapping algorithms underlying classical planning are subject to failure cases depending on the semantics of the scene and that this dependence is learnable: for instance, certain areas, objects or scene structures can be  reconstructed easier than others. We propose a hierarchical method composed of a high-level planner dynamically switching between a classical and a neural planner. We fully train all neural policies in simulation and evaluate the method in both simulation and real experiments with a LoCoBot robot, showing significant gains in performance, in particular in the real environment. We also qualitatively conjecture on the nature of data regularities exploited by the high-level planner.
\end{abstract}

\section{Introduction}

\noindent
Large-scale machine learning has had a significant impact on robotics, and in particular on navigation of mobile robots, where end-to-end training in simulated 3D environments like Habitat \cite{Savva_2019_ICCV} and AI-Thor \cite{AIThorKolve} has been proposed as an alternative to classical map and plan baselines. The potential advantages of learning to plan with high-capacity deep neural networks are the promise of complex decision functions, able to cope with large amounts of noise, sensor failure and unmodeled disturbances, and complex dependencies on scene semantics, which are difficult to design with handcrafted algorithms. This complexity comes with a price, the dependency on massive amounts of training data in the form of 3D scene models loaded into simulators. While the amount of data seen during training can be almost unlimited (modern models are trained on typically 100M --- up to 7B environment steps \cite{IsMappingNecessaryCVPR2022}), the main factors of variation are the number of scenes, which are limited due to the required effort of scanning physical buildings. Current datasets contain dozens or hundreds of scenes \cite{xia2018gibson,chang2018matterport3d}, with up to 1000 scenes for the latest HM3D dataset \cite{ramakrishnan2021hm3d}. Lack of sufficient diversity in scenes and the sim2real gap --- the difference between simulation and real environment --- limit the transfer of navigation performance to real environments.

For these reasons, classical map and plan baselines \cite{marder2010office,macenski2020marathon} are still competitive in many situations where the navigation task itself does not depend on complex high-level visual reasoning, and where maps can be estimated with sufficient reliability. In this work we ask two scientific questions: (1) are trained and classical planning strategies complementary and 
excel in different situations, and (2) can these different types of situations be clearly distinguished from visual observations, making it possible to exploit these regularities?

\begin{figure}[t]    \centering    
    \includegraphics[width=\linewidth]{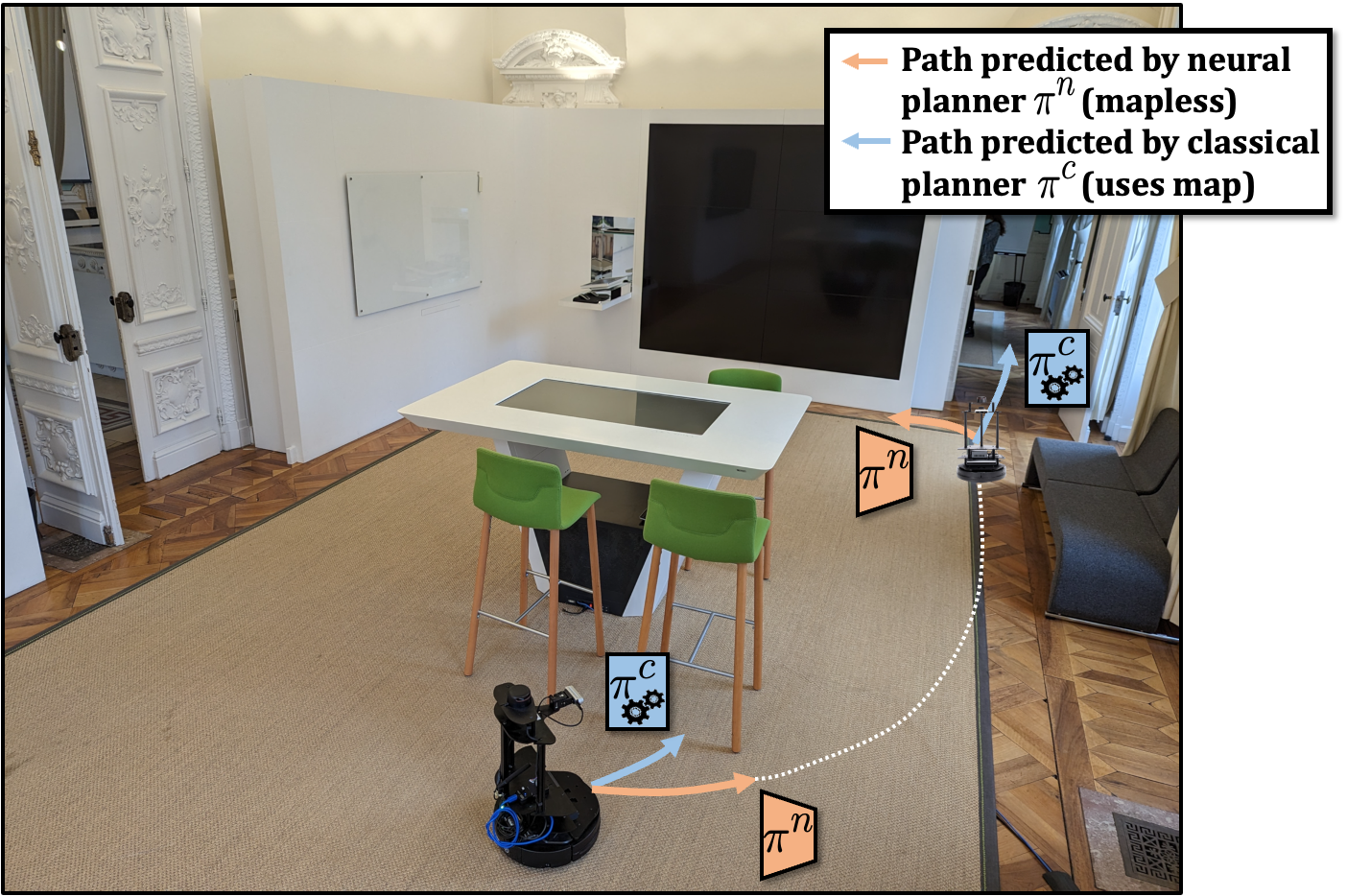}
    \caption{\label{fig:teaserscene}In indoor navigation problems, we present an agent which can resort to two different strategies, a \textcolor{ColNeural}{trained neural planner} and a \textcolor{ColClassical}{classical planner} based on occupancy maps. An additional \textcolor{ColHL}{high-level governor} is trained to switch between the two strategies based on learned regularities between planning performance and scene semantics, for instance that high chairs are not well reconstructed and lead to bad performance of a classical map-and-plan solution. We train in simulation and evaluate in, both, simulation and an office building using a real robot.}    
\end{figure}

\begin{figure*}[t] \centering
    \includegraphics[width=0.9\linewidth]{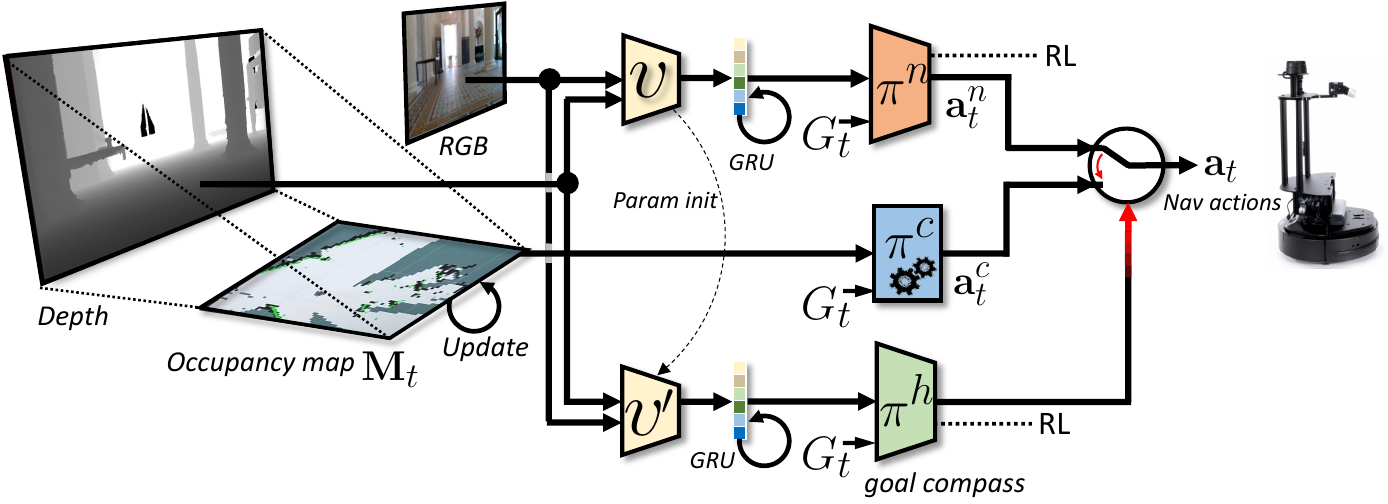}
    \captionof{figure}{\label{fig:schema}We distribute navigation decisions over two different planners: a \textcolor{ColNeural}{trained low-level planner $\pi^n$} takes RGB-D first-person input, and a \textcolor{ColClassical}{classical planner $\pi^c$} takes a metric occupancy map $\mathbf{M}_t$ as input. A \textcolor{ColHL}{high-level planner $\pi^h$} exploits regularities between scene elements and planning performance and learns to take a binary decision between these two planners, based only on first person inputs. The hidden state of the recurrent policy $\pi^n$ is updated even when the classical planner is used.     
    }
\end{figure*}

We explore these questions in a series of experiments and propose a new hybrid method combining classical and neural planning. Compared to existing hybrid solutions in the literature \cite{HybridDashoraLevine,Sim2RealStrategy,PRMRL2018}, our method is based on a trained combination. A high-level planner, trained with RL, dynamically switches between the two alternative planning methods and learns to adapts to the situation at hand, as shown in Fig. \ref{fig:schema}. To this end, it receives as input features extracted from first-person images, which may be useful to exploit correlations between scene semantics and planning performance. We also experiment with a variant which takes the high-level decision on, both, the first person input and occupancy map. The exact regularities picked up by the high-level planner may be complex, and we attempt to answer this question in the experimental part of this paper. To further motivate this approach beforehand, we mention possible scenarios: 3D scene structures difficult to reconstruct and to project into an occupancy map might be recognizable from their first person depth input, or linked to their semantic class and recognizable from the first person RGB input; 2D structures in the occupancy map harmful to classical or neural planning could be detectable directly; the trained low-level planner might be subject to biases picked up in simulation from spurious correlations, and these biases might be learnable by the high-level planner, switching over to classical planner when needed.

We claim the following contributions:
\begin{itemize}
\item a hybrid method switching between complementary navigation strategies based on a high-level planner trained with reinforcement learning on dense reward (geodesic distance to the goal).
\item Large-scale training in 3D photorealistic simulation using complex first person \mbox{RGB-D} input.
\item Transfer from simulation to a real environment and extensive experiments with a LoCoBot mobile robot.
\end{itemize}

\section{Related work}

\myParagraph{Navigation with mapping and planing} is the core capability of service robots since their introduction~\cite{burgard1998interactive}. Classic navigation stacks often assume  access to a pre-scanned map of the environment~\cite{burgard1998interactive,marder2010office,macenski2020marathon} and are composed of three main modules: mapping and localization using visual or Lidar SLAM~\cite{TBF2002probabilisticrobotics, labbe19rtabmap}, global planning with, for example, A*~\cite{konolige2000gradient} or Fast Marching Method (FMM)~\cite{sethian1996fast}, and low-level local path planning to reach intermediate waypoints \cite{fox1997dynamic,rosmann2015timed}. The classical planner used in this work does not have access to the environment map where it is deployed. It uses depth images and odometry to incrementally build a 2D egocentric occupancy map and localize the agent on it, while planning is done using FMM.

\myParagraph{End-to-End navigation}
directly trains an agent to predict actions from observed input, either with reinforcement learning (RL) or imitation learning (IL). Given the partial observable nature of the problem, the agent keeps latent memory, typically through a recurrent neural network. Additional structured neural memory has been proposed, e.g. neural metric maps \cite{Henriques_2018_CVPR,DBLP:conf/icpr/BeechingD0020}, semantic maps \cite{Chaplot20objectgoal}, neural 
topological maps \cite{Chaplot_2020_CVPR,savinov2018semiparametric,shah_viking_2022,beeching2020learning}
or implicit representations~\cite{ImplicitICLR2022,MarzaNERFArxiv2022}. Recently, it has also been proposed to replace recurrence by Transformers \cite{vaswani2017attention} with self-attention over the history of observations \cite{Fang_2019_CVPR,du_vtnet_2021,chen_think_2022,reed_generalist_2022}.  

\myParagraph{Modular and hybrid navigation}
Modular approaches decompose planning hierarchically. While the option framework \cite{Options1999} provides a generic solution in the context of planning with RL, specific solutions have been proposed for navigation. Typically, waypoints are proposed by a high-level (HL) planner, and then followed by a low-level (LL) planner. In one line of work, the HL planner is a trained model, which triggers actions by the LL planner, which is either also trained \cite{Chaplot2020Learning} or classically based on shortest path calculations on a map \cite{Chaplot20objectgoal,AssemICRA2023} or optimal control \cite{NavOptControlCORL2020}. In the complementary line of work, the HL planner is based on classical optimization based algorithms, e.g. Probabilistic Roadmaps~\cite{PRMRL2018} or shortest-path calculations in a high-level graph \cite{BeechingGamerland2022}. Both of these solutions defer point-to-point navigation to a LL planner trained with RL.

Hybrid methods combine classical planning with learned planning. Some of the modular approaches mentioned above can be considered to be hybrid, but there exist hybrid approaches in the literature which combine different planners more tightly and in a less modular way. 
In~\cite{Sim2RealStrategy} and similarly in~\cite{HybridDashoraLevine}, a planner trained with RL generates trajectories, which are used to generate a cost-map used by a classical planner. In \cite{PlayCatch2020}, a neural planner generates UAV trajectories which are then used by a model-predictive control as support for optimization. \textit{Neural-A*} learns a model predicting a cost-map for planning with a differentiable version of \textit{A*}, backpropagating a supervised loss through it \cite{neuralastar2021}. Similarly, \textit{Cognitive Mapping and Planning} \cite{gupta2017cognitive} learns a mapping function by backpropagating through a differentiable planner, in the form of \textit{Value Iteration Networks}~\cite{VINNips2016}. In~\cite{beeching2020learning}, a graph-network is imbued with inductive bias for planning with the~\textit{Bellman-Ford} algorithm. In \cite{MoleArxiv2022} both planners are neural, but one is blind.

All these existing solutions combine planners with different but handcrafted designs. In contrast, our method dynamically switches between types of planners with a trained model. Similar to our approach, in \cite{kastner2022all} a HL planner is trained on a schematic simulation to switch between a classic model-based planner and a learned planner for dynamic obstacle avoidance. However, this work considers a simple 2D set-up where all planners have access to the full map of the environment, perfect $360^{\circ}$ Lidar scans and exact obstacle positions, while our methods only access noisy first-person images in a realistic 3D simulator and a real robot. Also, the HL planner in~\cite{kastner2022all} tackles the considerably simpler task of selecting one of two options tailored to two different situations: efficiently navigate to a goal or avoid dynamic obstacles. In contrast, our HL planner has to learn to exploit subtle correlations between scene structure and semantics and planning performance to combine LL algorithms that are designed for the same task with comparable performance. 


\section{Learning to choose planners}
\label{sec:high-level}
\noindent
We address the \textit{PointGoal} task where an agent receives a visual observation $\mathbf{o}_t \in \mathbb{R}^{4{\times}H{\times}W}$ (an RGB-D image) and a Euclidean goal compass vector $G_t$ at each time step $t$ and must take actions $\mathbf{a}_t$ in a discrete action space $\Lambda = \{${\small \texttt{MOVE\_FORWARD 25cm}, \texttt{TURN\_LEFT} $10^{\circ}$, \texttt{TURN\_RIGHT} $10^{\circ}$ and \texttt{STOP}}$\}$. The \texttt{STOP} action terminates the episode successfully if the agent is within 0.2m of the goal, or unsuccessfully if not. 

As shown in Fig. \ref{fig:schema}, our method takes decisions at each time $t$ on whether to choose an action predicted by a neural planner $\pi^n$ or a classical planner $\pi^c$. We will first introduce each low-level planner and then the high-level governor $\pi^h$. In what follows, superscripts $.^n$,  $.^c$ and $.^h$ do not take numerical values but rather denote choices between the neural, classical, or high-level planner, respectively. Network architectures of all trainable functions will be provided in section \ref{sec:architectures}.

\subsection{The neural planner}
\label{ssec:neural}
\noindent 
The neural planner $\pi^n$ is trained in simulation to directly predict navigation actions $\mathbf{a}^n_t \in \Lambda$ from visual input $\mathbf{o}_t$. It sequentially builds a representation $\mathbf{h}_t$ from the sequence $\{\mathbf{o}_{t'}\}_{t'<t}$ of visual first-person observations, and then predicts a distribution over actions,
\begin{align}
\label{eq:agentupdate}
\mathbf{h}_t &= f^n(\mathbf{h}_{t-1}, v(\mathbf{o}_t), \mathbf{a}^n_{t-1}),  \\
p(\mathbf{a}^n_t) &= \pi^n(\mathbf{h}_t, G_t), 
\end{align}
where $f^n$ is the update function of a recurrent GRU network, with gates omitted from the notation for convenience; $v$ is a visual encoder, i.e. a trained ResNet extracting features from observations.

We train this planner end-to-end with PPO~\cite{schulman2017proximal} with the reward definition from~\cite{chattopadhyay2021robustnav},
\begin{equation}
r_t=\mathrm{K} \cdot \mathbb{I}_{\text {success }}-\Delta_t^{\mathrm{Geo}}-\lambda,
\label{eq:rewardn}
\end{equation}
where $K{=}2.5$, $\Delta_t^{\mathrm{Geo}}$ is the gain in geodesic distance to the goal, and slack cost $\lambda{=}0.01$ encourages efficiency.

\subsection{The classical planner}
\label{sec:classical}
\noindent
Numerous algorithms and implementations exist for planning based on dynamically estimated maps. We use the map and plan baseline approach proposed in \cite{gupta2017cognitive}, which maintains an egocentric metric occupancy map $\mathbf{M}_t \in [0,1]^{N{\times}M}$, called ``\textit{Egomap}'', over time by first inversely projecting the depth channel of the visual observation $\mathbf{o}_t$ (using intrinsics of the calibrated camera) and then pooling the resulting point cloud to the ground, resulting in a local bird's-eye-view map for this observation. Consecutive maps are aligned with odometry and integrated with max pooling, as in \cite{Chaplot20objectgoal}. Planning is performed on this map using FMM~\cite{sethian1996fast}.

The action space of a planner based on shortest path calculations is inherently tied to the underlying representation it uses for planning, which in our case is the resolution of the metric map $\mathbf{M}_t$: a navigation action is a part of a path in the graph structure of the map $\mathbf{M}_t$, i.e. the choice of an edge between two nodes. However, to align the action spaces of the two complementary navigation strategies, we chose to translate these predictions into actions taken from the discrete alphabet $\Lambda$ of the downstream navigation task. This not only facilitates the design of the high-level planner, but also allows to run both low-level planners simultaneously and maintain their respective states, as will be discussed in the next section. This translation is done with a well-known, publicly available, map and plan baseline\footnote{\url{https://github.com/s-gupta/map-plan-baseline}}.


\subsection{The high-level planner}
\label{sec:highlevel}
\noindent
The high-level planner $\pi^h$ takes a binary decision $d_t \in \{0,1\}$ on the choice of planners, such that the final navigation action $\mathbf{a}_t$ is given as 
\begin{equation}
\mathbf{a}_t = 
d_t \mathbf{a}^n_t + 
(1{-}d_t) \mathbf{a}^c_t. 
\end{equation}
The planner is implemented as a recurrent policy, which maintains a hidden state $\mathbf{r}_t$ with a GRU, denoted as $f^h$, and which takes as input features extracted from the first person input $\mathbf{o}_t$,
\begin{align}
\label{eq:agenthupdate}
\mathbf{r}_t &= f^h(\mathbf{r}_{t-1}, v'(\mathbf{o}_t), 
d_{t-1}),  
\\
p(d_t) &= \pi^h(\mathbf{r}_t, G_t). 
\end{align}
The feature extractor $v'$ has the same architecture as $v$ in Eq.~(\ref{eq:agentupdate}) 
and it is fine-tuned from the trained version of $v$.

The high-level planner is trained with PPO end-to-end, jointly with the encoder $v'$, with a reward 
used for the neural low-level planner in Eq.~(\ref{eq:rewardn}). We train with vectorized environments and maintain 12 agents per batch. The neural planner $\pi^n$ is operated in parallel to the classical one $\pi^c$, and its hidden state $\mathbf{h}_t$ is updated with Eq.~(\ref{eq:agentupdate}), even if it has \textit{not} been chosen by the high-level planner, by providing it with the action taken by $\pi^c$. This allows the neural planner to maintain a spatial internal representation during navigation consistent with what it experienced during training, regardless of its actual use in the hybrid setting. 
Two key design choices were necessary to make this possible: the alignment of the action spaces of the two planners (see Section \ref{sec:classical}), and the possibility of updating the internal state $\mathbf{h}$ with an action different from the one predicted by the agent $\pi^n$ itself. The latter is enabled through sampling actions stochastically from the predicted discrete distribution $p(\mathbf{a}^n_t)$ during training; this leads the agent to update its internal (spatial) representation of the scene not based on its previously predicted action, but on the effectively performed previous action $\mathbf{a}_{t-1}$ input to the policy in Eq.~(\ref{eq:agentupdate}). 

We add two remarks here. First, during training, we sample from the predicted distribution $p(\mathbf{a}^n_t)$, which is different from the distribution of frequent action choices by the competing classical planner $\pi^c$ --- we chose to ignore this difference. Second, as in large part of the literature, we train without actuation noise, i.e. the previous action $\mathbf{a}_{t-1}$ provides the exact odometry information during training. Previous work~\cite{kadian2020sim2real} shows that doing so improves performance in real-world experiments. At testing the learned policy is directly transferred to the noisy setting.

\subsection{Network architectures}
\label{sec:architectures}

\noindent
The visual encoders $v$ and $v'$ are ResNet18~\cite{he2016deep} architectures.
The recurrent policies are composed of GRUs $f^n$ and $f^h$ with $2$ layers and hidden states of size $512$. Previous actions $\mathbf{a}^n_{t-1}$ and goal compass vector $G_t$ are encoded with learned embeddings of size $32$.

\begin{figure}[t] \centering
\includegraphics[width=0.8\linewidth]{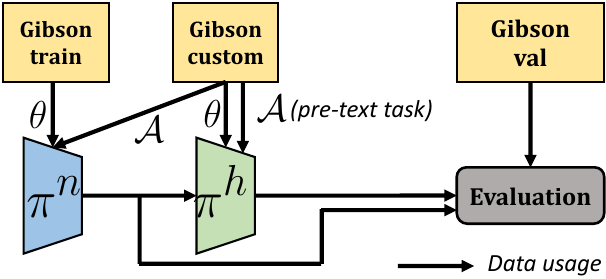}
\caption{\label{fig:split}\textbf{Training pipeline and data splits:} training the hybrid planner $\pi^h$ requires a custom data split, as training needs to be performed on data which have not been seen during training of the low-level neural planner $\pi^n$. ${\rightarrow}{\theta}$ indicates training network parameters with SGD training; ${\rightarrow}\mathcal{A}$ indicates architecture optimization (manual, through ``grad student descent''). We accepted some overlap in optimizing hyper-parameters, see the text.
However, \textit{evaluation was performed only on scenes unseen during training and hyper-parameter optimization}.
}
\end{figure}

\section{Experimental Results}
\label{sec:experiments}

\myParagraph{Experimental setup} 
all training was performed in simulation only with the Habitat simulator~\cite{Savva_2019_ICCV} and scenes from the \emph{Gibson} dataset~\cite{xia2018gibson}, which contains 3.6M episodes over 72 different scenes for training, and 994 navigation episodes over 14 scenes for validation. We test the system in both simulation, with additional noise, and a real physical robot, in particular a \textit{LoCoBot} robot~\cite{locobot} [ \includegraphics[height=4mm]{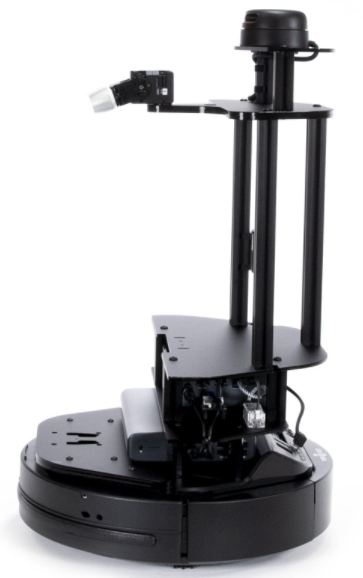} ] equipped with an Intel~\textit{RealSense} D435i RGB-D camera and a single-ray Lidar of type \textit{RPLIDAR A2M8}. 

The agent receives an RGB-D observation of size 160${\times}$120 pixels at each step, which in simulation matches the extrinsic and intrinsic parameters (position, the field of view and aspect ratio) of the onboard camera on the LoCoBot. It also gets a goal compass vector in the form of the Euclidean distance and direction, provided as privileged information in the simulator, and by the robot's position estimation system in experiments with the real physical robot. In the robot experiments this is done using the default \texttt{ROS} implementation of the Adaptive Monte-Carlo Localization algorithm~\cite{TBF2002probabilisticrobotics}, which is based on RTAB-Map, a 2D metric representation~\cite{labbe19rtabmap} generated from Lidar input. The Lidar is only used to localize the robot, while sensing, mapping and planning are based solely on RGB-D camera input.

\myParagraph{Simulator settings} we removed the possibility of the robot to slide across the walls (\textit{sliding OFF}). This makes the PointGoal navigation task more challenging for both low-level planners, and previous work~\cite{kadian2020sim2real} finds this setting crucial for real-world deployment. We configure the Habitat simulator and adjusted it to the properties of the physical robot (LoCoBot) and its sensors: FOV of 56° camera, frame size of 160{$\times$}120 and a compatible camera position. For the experiments which involved evaluation in simulation, i.e. Tables~\ref{tab:lowlevelsim},~\ref{tab:simresults} and~\ref{tab:ablationsinput}, we used a second simulator configuration which is compatible with the prior work~\cite{DBLP:conf/nips/WaniPJCS20,MarzaIROS2022}. It includes a FOV of 79° and camera frames of size 256{$\times$}256. 

\myParagraph{Data splits and training pipeline} 
as usually done in the relevant literature, we report results on the validation set of the Gibson dataset, as the test set is not available. However, to obtain optimal performance, this requires additional splits for validating the different models (hyperparameter optimization). In our case, differently from the classical settings, we require additional splits due to the fact that the high-level planner is trained on output of the neural low-level planner. Therefore, the high-level planner $\pi^h$ needs to be trained on data different from training $\pi^n$, in order to avoid a potential bias of $\pi^h$ trained on an overconfident $\pi^n$ overfitting on its training environment and leading to skewed decisions. We therefore introduced an additional dataset split called \textit{Gibson-custom} which consists of 1036 episodes over 14 unused scenes selected from the full Gibson dataset.

Figure~\ref{fig:split} illustrates the training pipeline. The neural low-level planner $\pi^n$ is trained on the Gibson training set. The high-level planner is trained on the custom split, and the Gibson validation set is used to report results. The hyper-parameters (network architectures $\mathcal{A}$) of the low-level planner $\pi^n$ have been optimized on \textit{Gibson-custom}. In other words, we accepted a small possibility of training $\pi^h$ on overconfident decisions based on validation overfit, but we judged this risk to be small. To work around the requirement of one more data split to optimize the hyper-parameters of the high-level planner, we optimized them using a proxy task, namely exploration. More precisely, we use the network architecture of the high-level planner in~\cite{AssemICRA2023}. This planner provides high-level decisions of different nature, waypoint coordinates followed by a low-level planner, and we adapted its later layers to take binary decisions instead. These decisions did not interfere with the soundness of the evaluation protocol: \textit{all evaluation was performed only on scenes unseen during training or hyper-parameter optimization}.

\begin{table}[t] \centering
{\small
\setlength{\aboverulesep}{0pt}
\setlength{\belowrulesep}{0pt}
\begin{tabular}{ a b b b c c}
\toprule
 \textbf{Agent} & \textbf{Input} & 
 \textbf{Train-$\mathcal{N}$} & \textbf{Test-$\mathcal{N}$} &
 \textbf{Succ.} & \textbf{SPL}  \\
\midrule
\textbf{Neural} & RGB-D & 
\xmark&\xmark&
$90.94$ & 
$77.14$ 
\\
\textbf{Neural} & RGB-D & 
\xmark&Redwood+&
$87.87$ & 
$74.21$ 
\\
\textbf{Neural} & RGB-D & 
Redwood+ & Redwood+&
$89.24$ & 
$75.92$ 
\\
\textbf{Classical} & Egomap  & 
N/A&\xmark&
$87.93$ & 
$79.69$  
\\
\textbf{Classical} & Egomap  & 
N/A & Redwood &
$78.67$ & 
$72.17$ 
\\
\bottomrule
\end{tabular}
}
\caption{\label{tab:lowlevelsim}Performance of different low-level planners in simulation (Gibson-val), where $\mathcal{N}$ is the noise model. The table shows how the difference between Redwood and Redwood+ impacts the neural planner.}
\end{table}

\begin{table}[t] \centering
{\small
\setlength{\aboverulesep}{0pt}
\setlength{\belowrulesep}{0pt}
\begin{tabular}{ a l l}
\toprule
 \textbf{Agent}         &  \textbf{Success} & \textbf{SPL}  \\
\midrule
\textbf{Classical only ($\pi^c$)}  & 
$78.67$ & 
$72.17$ 
\\
\textbf{Neural only ($\pi^n$)}  & 
$89.24$ & 
$75.92$ 
\\
\midrule
\textbf{Random HL-decisions}  & 
$88.88${\scriptsize $\pm 1.4$} & 
$73.78${\scriptsize $\pm 1.1$} 
\\
\textbf{Hybrid (Ours)} & 
$90.64$ & 
$75.62$ 
\\
\bottomrule
\end{tabular}
}
\caption{\label{tab:simresults}\textbf{Performance of the hybrid method in simulation}, tested with Redwood+ Noise.}
\end{table}

\begin{table}[t] \centering
{\small
\setlength{\aboverulesep}{0pt}
\setlength{\belowrulesep}{0pt}
\begin{tabular}{ddcc}
\toprule
 \multicolumn{2}{d}{\textbf{--- Input to $\pi^h$ ---}} & 
 \textbf{Success} & \textbf{SPL}  
 \\
 {\footnotesize $1^{st}$ person}  &
 {\footnotesize Egomap$^{*}$}  &&
 \\
\midrule
\textbf{RGB-D} & \xmark  & 
$90.64$ & 
$75.62$ 
\\
\midrule
\textbf{RGB-D} & \checkmark & 
$90.85$ & 
$75.78$ 
\\
\xmark & \checkmark & 
$89.03$ & 
$74.87$ 
\\

\bottomrule
\end{tabular}
}
\caption{\label{tab:ablationsinput}\textbf{Impact of the privileged map information} on the high-level planner: simulation with Noise on Gibson-val.}
\end{table}

\begin{table}[t] \centering
{\small
\setlength{\aboverulesep}{0pt}
\setlength{\belowrulesep}{0pt}
\begin{tabular}{ a c c c}
\toprule
 \textbf{Agent}         &  \textbf{Success} & \textbf{SPL} & \textbf{SPL}$^{Succ}$ \\
\midrule
\textbf{Classical only ($\pi^c$)}  & 
$33.33$ & 
$27.19$ &
$81.57$
\\
\textbf{Neural only ($\pi^n$)}  & 
$100.00$ & 
$58.55$ &
$58.55$ 
\\
\textbf{Hybrid (Ours)} & 
$100.00$ & 
$72.50$  &
$72.50$ 
\\
\bottomrule
\end{tabular}
}
\caption{\label{tab:realresults}\textbf{Performance of the hybrid method in the real environment}: A LoCoBot in a real classical European office building, on 12 test episodes. SPL$^{Succ}$ indicates the SPL metric only for the episodes which were succeeded.}
\end{table}

\subsection{Quantitative Results}
\label{ssec:quantitative}

\myParagraph{Performance of the low-level planners} we evaluated the two low-level planners in simulation and report results in Table~\ref{tab:lowlevelsim} in terms of Success and Success weighted by (normalized inverse) Path Length (SPL)~\cite{DBLP:journals/corr/abs-1807-06757}. We explored different noise types on the depth observation, which is used, both, as input to the neural planner and to generate the Egomap for the classical planner. Redwood noise \cite{Teichman-RSS-13} is often used in evaluation of navigation, and we also explored a variant which we call ``\textit{Redwood+}'' in Table~\ref{tab:lowlevelsim}. It is motivated by the observation that in the standard Habitat implementation of the depth noise model, a depth {\tt D} above a given threshold {\tt T} was set to zero\footnote{\scriptsize\url{https://github.com/facebookresearch/habitat-sim/blob/d3d150c62f7d47c4350dd64d798017b2f47e66a9/habitat_sim/sensors/noise_models/redwood_depth_noise_model.py\#L73}}, i.e.
{\small \verb!if(D>T)D=0!},
which is the inverse behavior of the noiseless setting, which truncates depth, ie. {\small \verb!if(D>T)D=T!}.
We argue that this extremely strong discrepancy does not fall into the category of noise but rather to a change in the nature of the sensor (it corresponds to the behavior of certain depth sensors like Kinect), degrades transfer and does not allow a sound evaluation; we therefore replaced this zeroing version with the truncating variant. This difference mostly has an impact on the neural planner, not the classical one.

As we can see in Table~\ref{tab:lowlevelsim}, the planners perform similarly in the noiseless environment. However, the classical planner's performance drops significantly in the noisy environment, due to a degraded quality of the Egomap on which planning is performed. The impact of noise on the neural planner is less pronounced.

\begin{figure*}[t] \centering
\includegraphics[width=0.85\linewidth]{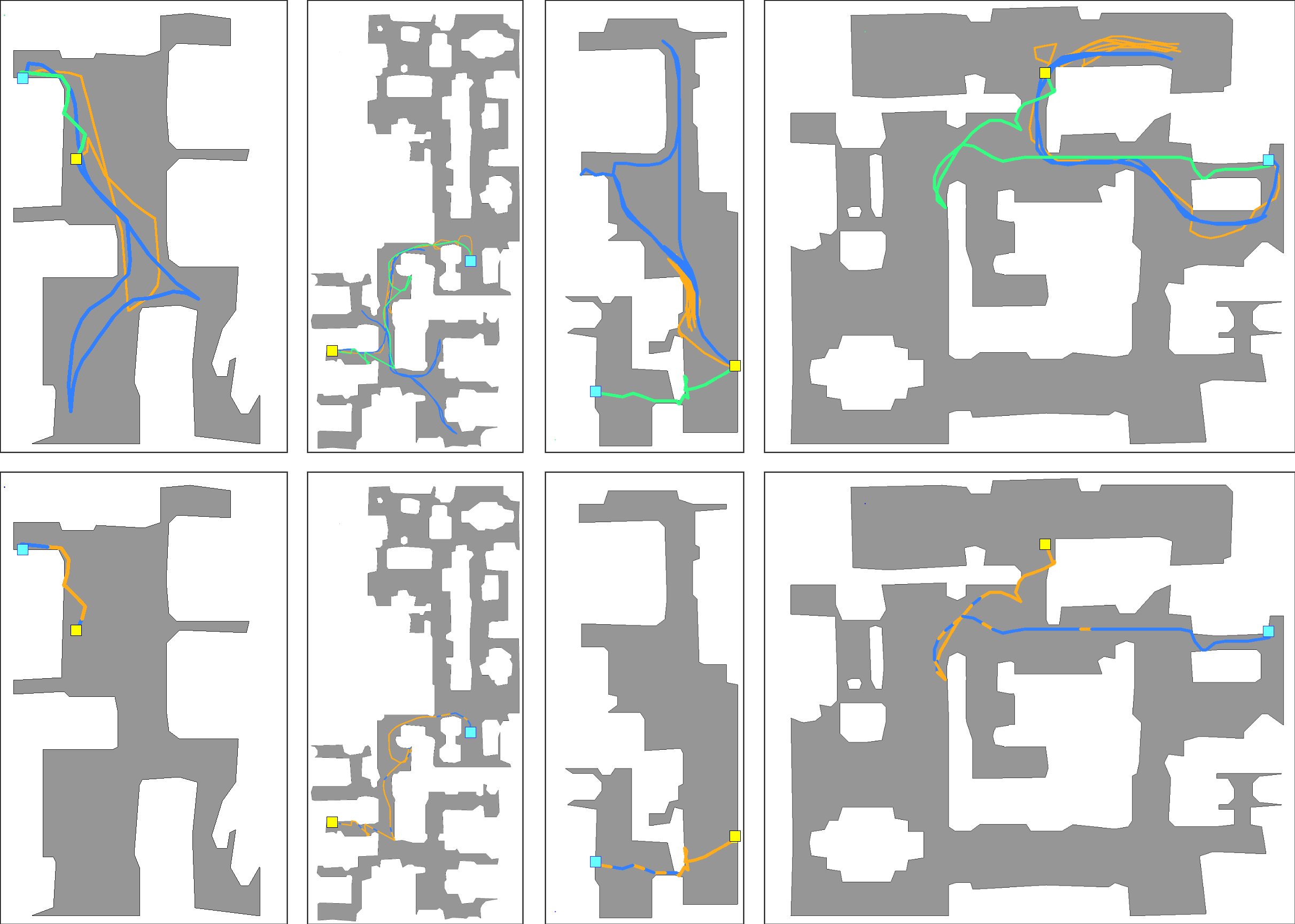}
\caption{\label{fig:sample_traj} \textbf{Rollouts of four episodes in different environments}: the robot starts at \textcolor{yellow}{$\blacksquare$} and has to reach the goal position \textcolor{cyan}{$\blacksquare$}. \textbf{Top row}: comparing trajectories taken by the \textcolor{ColNeural}{trained neural planner}, \textcolor{ColClassical}{classical planner} and our \textcolor{ColHL}{hybrid planner}. \textbf{Bottom row}: each step of the hybrid planner path in the top row is colored with the chosen low-level planner, \textcolor{ColNeural}{neural} or \textcolor{ColClassical}{classical}.
}
\vspace{-0.1cm}
\end{figure*}

\begin{figure}[t] \centering
\includegraphics[width=0.99\linewidth]{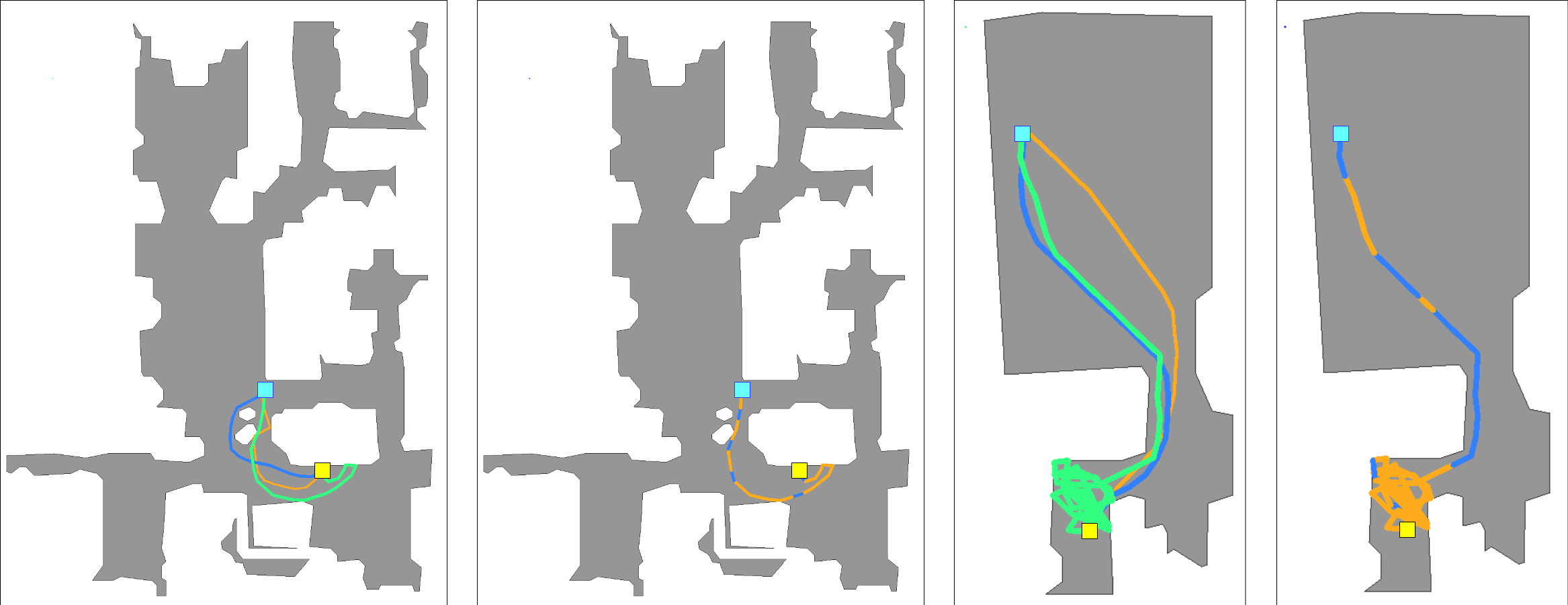}
\caption{\label{fig:fail} \textbf{Failure cases}: Two examples where the \textcolor{ColHL}{hybrid planner} perform worse than the \textcolor{ColNeural}{neural} and \textcolor{ColClassical}{classical} planners.
}\vspace{-0.1cm}
\end{figure}

\begin{figure*}[t] \centering
\includegraphics[width=\linewidth]{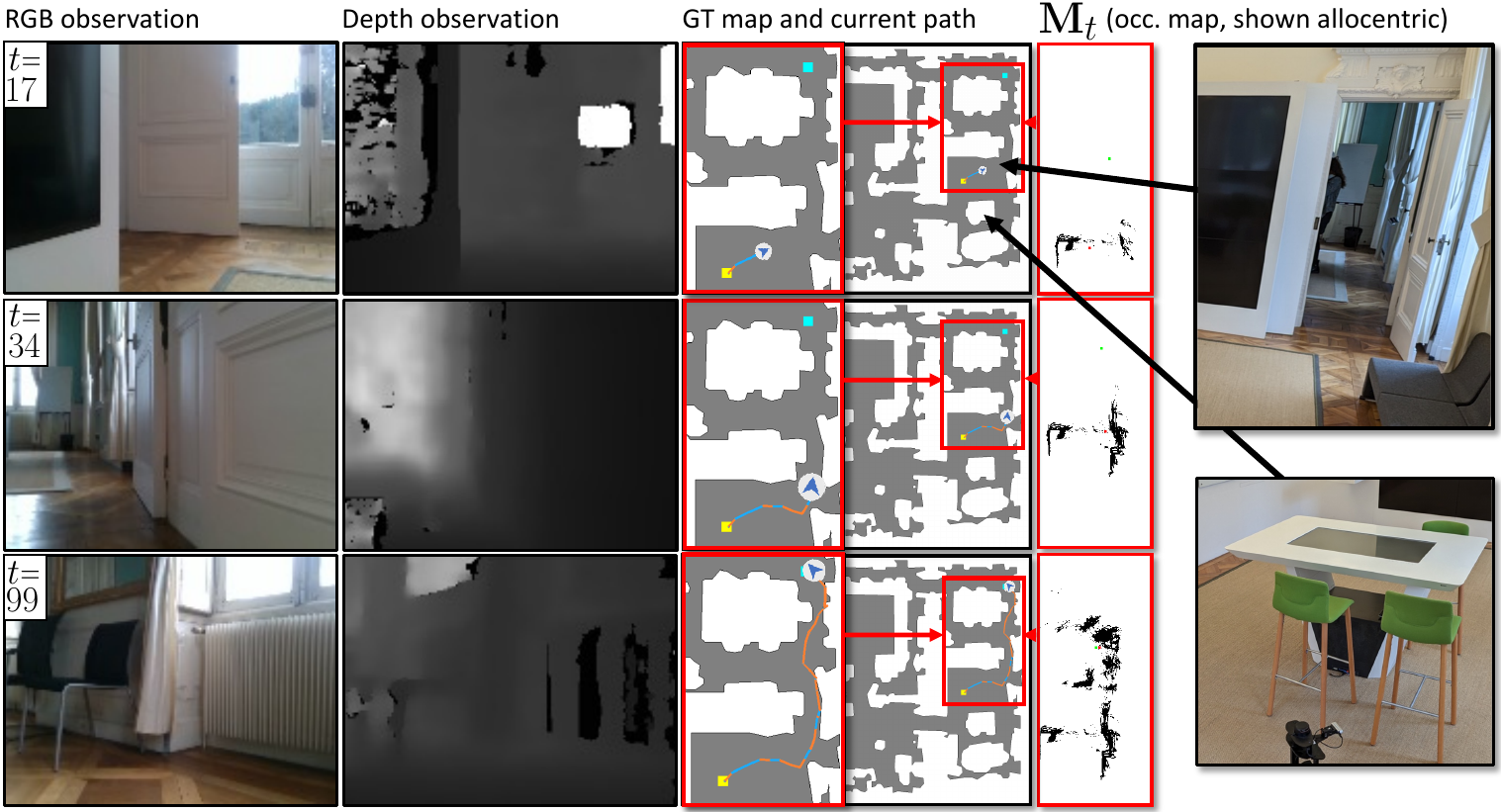}
\caption{\label{fig:rollouts_traj}A rollout of an episode, showing inputs and representations. We overlay the current path over the ground-truth (GT) map, color coding \textcolor{ColNeural}{neural steps} and \textcolor{ColClassical}{classical steps}. The robot starts at \textcolor{yellow}{$\blacksquare$} and has to reach the goal position \textcolor{cyan}{$\blacksquare$}. For better comparability, the Egomap $M_t$ is shown here not as an Egomap but in an allocentric frame. The big black arrows indicate parts of the map corresponding to the scene shown in Figure \ref{fig:teaserscene}.}
\end{figure*}

\myParagraph{Hybrid planning in simulation} Table~\ref{tab:simresults} compares performances of the low-level planners with the proposed hybrid planner. The hybrid planner outperforms both low-level variants in the Success rate, and also outperforms the baseline random high-level decisions. This version of the HL-planner takes as input the first-person RGB-D observation and thus exploits regularities between the currently observed scene structure and low-level planning performance. 

We also explored whether there exist correlations between the 2D structure of the occupancy map and performance of the two low-level planners and a high-level planner on this input, additional to first-person input. As a proof of concept, and to minimize the impact of noise and purely focus on scene structure, we performed this experiment with a noiseless Egomap$^{*}$ generated through privileged information in the simulator. Results in Table~\ref{tab:ablationsinput} show that the impact of the scene structure is minimal.

\myParagraph{Experiments with a real robot} we carried out 12 episodes with the LoCoBot in a classical European office building (see Figure \ref{fig:teaserscene}) with multiple rooms and challenging situations, like thick carpets and multiple big windows that pose problems to the onboard depth sensors. We report results in Table~\ref{tab:realresults} shows that the hybrid solution outperforms both individual standalone low-level planners significantly. 

The low performance of the classical planner is explained by the fact that the maps it produces are noisy. One particular aspect we can single out is the choice of map integration over time as in~\cite{Chaplot20objectgoal}, which uses max pooling to combine the bird's-eye-view estimate of the latest observation with the current global bird's-eye-view map. This choice is simple to implement but not as robust as state-of-the art Lidar based solutions like \mbox{RTAB-Map}, that feature a sophisticate probabilistic model and loop closure. Our choice is motivated by the objective to minimize the algorithmic sim2real gap of the two representations: the current state-of-the-art mapping solutions are difficult to integrate into a simulator like Habitat. The goal of these experiences is not to achieve state-of-the-art performance in planning, but to study the possibility of learning regularities in planning performance.

Another reason of the low classical planner's performance is the lack of high-level reasoning in case of missing information. The planning algorithm assumes that any unobserved area in the map is navigable, it corrects these estimates when a new observation becomes available and re-plans. This leads to backtracking and long trajectories. In contrast, the neural planner takes decisions not based on a 2D occupancy map but on first person input, which provides better cues on dead ends. It can also learn higher-level visual reasoning from a large amount of environment interactions and can avoid situations where backtracking would be needed otherwise. To quantify this behavior, in Table~\ref{tab:realresults} we also provide an additional metric, SPL$^{Succ}$, which corresponds to the SPL metric only for the successful episodes by the respective planner. This metric is high for the classical planner, which is efficient in cases where it does not get lost in local minima and requires extensive backtracking, leading to exceeding the maximum number of steps the task allows (=500).

The hybrid planner achieves the same 100\% success rate as the standalone neural planner, but with a better SPL metric (72.50 instead of 58.55), which indicates that it is more efficient. The neural planner indeed spends more time exploring, which makes it more robust than the classical planner in certain situations but can also be harmful in others. The hybrid planner manages to combine both advantages by dynamically switching between them.

\subsection{Qualitative Results}
\label{ssec:qualitative}

\myParagraph{Sample trajectories in simulation} Figure~\ref{fig:sample_traj} shows four episodes in different environments. The top row of pictures compares the behaviour of the neural, classical and hybrid planners. The bottom row shows the decisions taken by the hybrid planner in each episode. Our hybrid solution combines the low-level planners to solve long-horizon navigation tasks by exploring complex unknown environments, maneuvering in narrow spaces and efficiently reaching the goal. The hybrid planner starts episodes by using mainly the neural planner, which has better exploration capabilities. The neural planner is the preferred choice when the robot has to pass through a narrow corridor, as in the 3$^{\textrm{rd}}$ example. The classical planner is frequently employed towards the end of the episodes, when the path to the goal is clearer, as in the 4$^{\textrm{th}}$ example.

While the proposed hybrid approach has on average better navigation performance, this strategy can occasionally perform worse than the individual low-level planners. Figure~\ref{fig:fail} shows two typical failure cases: on the left, the hybrid planner selects the neural planner to start the episode, but it explores the wrong side of the scene, so the hybrid planner has to take a long detour to reach the goal. On the right, a more rare but dramatic failure case occurs when the hybrid planner, driven mainly by the neural planner, gets lost and starts to frenetically explore the environment. We conjecture that this might be due to few actions executed by the classical planner that put the neural planner in an unstable state.

\myParagraph{Example robot rollout} Figure~\ref{fig:rollouts_traj} shows an example episode rollout for three time instants $t=17,34,99$, including the first person input $\mathbf{o}_t$ (RGB and depth), the GT map with the overlaid path and color coded high-level decisions, as well as the occupancy Egomap $\mathbf{M}_t$ --- which we display in an allocentric way (and not as an egocentric map) for better comparability with the GT map. During the episode, we can notice that the HL planner relies more on the neural planner, which is more capable of navigating through narrow spaces encountered in this episode, except when the robot deviates from the most promising direction (towards the door) and the classical planner is chosen. Indeed, until $t=17$, the classical planner dominate the HL decisions, and guides the robot towards the goal. After a segment where the neural planner is chosen, at $t=34$, the classical planner takes over again to readjust the direction of the robot; then, until the end of the episode, the HL planner switches again to the neural one to traverse the final narrow passage.

\section{Conclusion}
\label{sec:conclusion}
\noindent
We have presented a hybrid method for navigation in real environments, which combines advantages of classical planning methods based on occupancy maps and shortest path computations with the power of neural methods trained in large-scale 3D photo-realistic simulations. We used RL to train a neural HL planner to dynamically switch between the two different LL planners and showed that they are complementary. Our experiments provide evidence for correlations between the observed scene structure and the difference in planning performance between the two LL planners, which are exploited by the hybrid solution. We have evaluated the proposed method in, both, simulation and a robot in a real office building, showing that the learned regularities transfer well. Future work will focus on learning the high-level decision on real data in the form of offline trajectories captured with a physical robot.

\myParagraph{Acknowledgement} We thank ANR for support through AI-chair grant ``Remember'' (ANR-20-CHIA-0018).


\begin{thebibliography}{10}
    \providecommand{\url}[1]{#1}
    \csname url@rmstyle\endcsname
    \providecommand{\newblock}{\relax}
    \providecommand{\bibinfo}[2]{#2}
    \providecommand\BIBentrySTDinterwordspacing{\spaceskip=0pt\relax}
    \providecommand\BIBentryALTinterwordstretchfactor{4}
    \providecommand\BIBentryALTinterwordspacing{\spaceskip=\fontdimen2\font plus
    \BIBentryALTinterwordstretchfactor\fontdimen3\font minus
      \fontdimen4\font\relax}
    \providecommand\BIBforeignlanguage[2]{{%
    \expandafter\ifx\csname l@#1\endcsname\relax
    \typeout{** WARNING: IEEEtran.bst: No hyphenation pattern has been}%
    \typeout{** loaded for the language `#1'. Using the pattern for}%
    \typeout{** the default language instead.}%
    \else
    \language=\csname l@#1\endcsname
    \fi
    #2}}
    
    \bibitem{Savva_2019_ICCV}
    M.~Savva, A.~Kadian, O.~Maksymets, Y.~Zhao, E.~Wijmans, B.~Jain, J.~Straub,
      J.~Liu, V.~Koltun, J.~Malik, D.~Parikh, and D.~Batra, ``Habitat: A platform
      for embodied ai research,'' in \emph{ICCV}, 2019.
    
    \bibitem{AIThorKolve}
    E.~Kolve, R.~Mottaghi, D.~Gordon, Y.~Zhu, A.~Gupta, and A.~Farhadi,
      ``{AI2-THOR:} an interactive 3d environment for visual {AI},''
      \emph{arxiv:1712.05474}, 2017.
    
    \bibitem{IsMappingNecessaryCVPR2022}
    R.~Partsey, E.~Wijmans, N.~Yokoyama, O.~Dobosevych, D.~Batra, and O.~Maksymets,
      ``Is mapping necessary for realistic pointgoal navigation?'' in \emph{CVPR},
      2022.
    
    \bibitem{xia2018gibson}
    F.~Xia, A.~R. Zamir, Z.~He, A.~Sax, J.~Malik, and S.~Savarese, ``Gibson env:
      Real-world perception for embodied agents,'' in \emph{CVPR}, 2018.
    
    \bibitem{chang2018matterport3d}
    A.~Chang, A.~Dai, T.~Funkhouser, M.~Halber, M.~Niebner, M.~Savva, S.~Song,
      A.~Zeng, and Y.~Zhang, ``Matterport{3D}: Learning from {RGB-D} data in indoor
      environments,'' in \emph{3DVision}, 2018.
    
    \bibitem{ramakrishnan2021hm3d}
    S.~K. Ramakrishnan, A.~Gokaslan, E.~Wijmans, O.~Maksymets, A.~Clegg, J.~M.
      Turner, E.~Undersander, W.~Galuba, A.~Westbury, A.~X. Chang, M.~Savva,
      Y.~Zhao, and D.~Batra, ``Habitat-matterport {3D} dataset ({HM3D}): 1000
      large-scale {3D} environments for embodied {AI},'' in \emph{NeurIPS ---
      Datasets and Benchmarks Track}, 2021.
    
    \bibitem{marder2010office}
    E.~Marder-Eppstein, E.~Berger, T.~Foote, B.~Gerkey, and K.~Konolige, ``The
      office marathon: Robust navigation in an indoor office environment,'' in
      \emph{ICRA}, 2010.
    
    \bibitem{macenski2020marathon}
    S.~Macenski, F.~Mart{\'\i}n, R.~White, and J.~G. Clavero, ``The marathon 2: A
      navigation system,'' in \emph{IROS}, 2020.
    
    \bibitem{HybridDashoraLevine}
    N.~Dashora, D.~Shin, D.~Shah, H.~Leopold, D.~Fan, A.~Agha-Mohammadi,
      N.~Rhinehart, and S.~Levine, ``Hybrid imitative planning with geometric and
      predictive costs in off-road environments,'' in \emph{NeurIPS Workshop on
      Deep-RL}, 2021.
    
    \bibitem{Sim2RealStrategy}
    K.~Weerakoon, A.~Sathyamoorthy, and D.~Manocha, ``Sim-to-real strategy for
      spatially aware robot navigation in uneven outdoor environments,'' in
      \emph{ICRA Workshop Releasing Robots into the Wild}, 2022.
    
    \bibitem{PRMRL2018}
    {A. Faust, O. Ramirez, M. Fiser, K. Oslund, A. Francis, J. Davidson, L. Tapia},
      ``{PRM-RL}: Long-range robotic navigation tasks by combining reinforcement
      learning and sampling-based planning,'' in \emph{ICRA}, 2018.
    
    \bibitem{burgard1998interactive}
    W.~Burgard, A.~B. Cremers, D.~Fox, D.~H{\"a}hnel, G.~Lakemeyer, D.~Schulz,
      W.~Steiner, and S.~Thrun, ``The interactive museum tour-guide robot,'' in
      \emph{Aaai/iaai}, 1998, pp. 11--18.
    
    \bibitem{TBF2002probabilisticrobotics}
    S.~Thrun, W.~Burgard, and D.~Fox, \emph{Probabilistic Robotics}.\hskip 1em plus
      0.5em minus 0.4em\relax MIT Press, 2005.
    
    \bibitem{labbe19rtabmap}
    M.~Labbé and F.~Michaud, ``{RTAB-Map} as an open-source lidar and visual
      simultaneous localization and mapping library for large-scale and long-term
      online operation,'' \emph{Journal of Field Robotics}, vol.~36, no.~2, pp.
      416--446, 2019.
    
    \bibitem{konolige2000gradient}
    K.~Konolige, ``A gradient method for realtime robot control,'' in \emph{IROS},
      2000.
    
    \bibitem{sethian1996fast}
    J.~A. Sethian, ``A fast marching level set method for monotonically advancing
      fronts.'' \emph{PNAS}, vol.~93, no.~4, pp. 1591--1595, 1996.
    
    \bibitem{fox1997dynamic}
    D.~Fox, W.~Burgard, and S.~Thrun, ``The dynamic window approach to collision
      avoidance,'' \emph{IEEE Rob. \& Aut. Magazine}, 1997.
    
    \bibitem{rosmann2015timed}
    C.~R{\"o}smann, F.~Hoffmann, and T.~Bertram, ``Timed-elastic-bands for
      time-optimal point-to-point nonlinear model predictive control,'' in
      \emph{European Control Conference (ECC)}, 2015.
    
    \bibitem{Henriques_2018_CVPR}
    J.~F. Henriques and A.~Vedaldi, ``Mapnet: An allocentric spatial memory for
      mapping environments,'' in \emph{CVPR}, 2018.
    
    \bibitem{DBLP:conf/icpr/BeechingD0020}
    E.~Beeching, J.~Dibangoye, O.~Simonin, and C.~Wolf, ``{Deep Reinforcement
      Learning on a Budget: 3D Control and Reasoning without a Supercomputer},'' in
      \emph{ICPR}, 2020.
    
    \bibitem{Chaplot20objectgoal}
    D.~S. Chaplot, D.~Gandhi, A.~Gupta, and R.~Salakhutdinov, ``Object goal
      navigation using goal-oriented semantic expl.'' in \emph{{NeurIPS}}, 2020.
    
    \bibitem{Chaplot_2020_CVPR}
    D.~S. Chaplot, R.~Salakhutdinov, A.~Gupta, and S.~Gupta, ``Neural topological
      {SLAM} for visual navigation,'' in \emph{CVPR}, 2020.
    
    \bibitem{savinov2018semiparametric}
    N.~Savinov, A.~Dosovitskiy, and V.~Koltun, ``Semi-parametric topological memory
      for navigation,'' in \emph{ICLR}, 2018.
    
    \bibitem{shah_viking_2022}
    D.~Shah and S.~Levine, ``{ViKiNG: Vision-Based Kilometer-Scale Navigation with
      Geographic Hints},'' in \emph{RSS}, 2022.
    
    \bibitem{beeching2020learning}
    {E. Beeching and J. Dibangoye and O. Simonin and C. Wolf}, ``Learning to plan
      with uncertain topological maps,'' in \emph{ECCV}, 2020.
    
    \bibitem{ImplicitICLR2022}
    X.~Li, S.~D. Mello, X.~Wang, J.~K. M.-H.~Yang, and S.~Liu, ``Learning
      continuous environment fields via implicit functions,'' in \emph{ICLR}, 2022.
    
    \bibitem{MarzaNERFArxiv2022}
    P.~Marza, L.~Matignon, O.~Simonin, and C.~Wolf, ``{Multi-Object Navigation with
      dynamically learned neural implicit representations.}'' in \emph{ICCV}, 2023.
    
    \bibitem{vaswani2017attention}
    A.~Vaswani, N.~Shazeer, N.~Parmar, J.~Uszkoreit, L.~Jones, A.~N. Gomez,
      {\L}.~Kaiser, and I.~Polosukhin, ``Attention is all you need,'' in
      \emph{NeurIPS}, 2017.
    
    \bibitem{Fang_2019_CVPR}
    K.~Fang, A.~Toshev, L.~Fei-Fei, and S.~Savarese, ``Scene memory transformer for
      embodied agents in long-horizon tasks,'' in \emph{CVPR}, 2019.
    
    \bibitem{du_vtnet_2021}
    H.~Du, X.~Yu, and L.~Zheng, ``{VTNet}: {Visual} {Transformer} {Network} for
      {Object} {Goal} {Navigation},'' in \emph{{ICLR}}, 2021.
    
    \bibitem{chen_think_2022}
    S.~Chen, P.-L. Guhur, M.~Tapaswi, C.~Schmid, and I.~Laptev, ``Think {Global},
      {Act} {Local}: {Dual}-scale {Graph} {Transformer} for {Vision}-and-{Language}
      {Navigation},'' in \emph{{CVPR}}, 2022.
    
    \bibitem{reed_generalist_2022}
    S.~Reed, K.~Zolna, E.~Parisotto, S.~G. Colmenarejo, A.~Novikov, G.~Barth-Maron,
      M.~Gimenez, Y.~Sulsky, J.~Kay, J.~T. Springenberg, T.~Eccles, J.~Bruce,
      A.~Razavi, A.~Edwards, N.~Heess, Y.~Chen, R.~Hadsell, O.~Vinyals, M.~Bordbar,
      and N.~de~Freitas, ``A {Generalist} {Agent},'' \emph{arXiv:2205.06175}, 2022.
    
    \bibitem{Options1999}
    R.~Sutton, D.~Precup, and S.~Singh, ``Between {MDPs} and semi-{MDPs}: A
      framework for temporal abstraction in reinforcement learning,''
      \emph{Artificial Intelligence}, vol. 112, pp. 181--211, 1999.
    
    \bibitem{Chaplot2020Learning}
    D.~S. Chaplot, D.~Gandhi, S.~Gupta, A.~Gupta, and R.~Salakhutdinov, ``Learning
      to explore using active neural {SLAM},'' in \emph{ICLR}, 2020.
    
    \bibitem{AssemICRA2023}
    A.~Sadek, G.~Bono, B.~Chidlovskii, A.~Baskurt, and C.~Wolf, ``{Multi-Object
      Navigation in real environments using hybrid policies},'' in \emph{{ICRA}},
      2023.
    
    \bibitem{NavOptControlCORL2020}
    S.~Bansal, V.~Tolani, S.~Gupta, J.~Malik, and C.~Tomlin, ``{Combining Optimal
      Control and Learning for Visual Navigation in Novel Environments},'' in
      \emph{{CORL}}, 2020.
    
    \bibitem{BeechingGamerland2022}
    E.~Beeching, M.~Peter, P.~Marcotte, J.~Dibangoye, O.~Simonin, J.~Romoff, and
      C.~Wolf, ``{Graph augmented Deep Reinforcement Learning in the GameRLand3D
      environment},'' in \emph{{AAAI Workshop on Reinforcement Learning in Games}},
      2022.
    
    \bibitem{PlayCatch2020}
    K.-H. Zeng, R.~Mottaghi, L.~Weihs, and A.~Farhadi, ``Visual {Reaction}:
      {Learning} to {Play} {Catch} with {Your} {Drone},'' in \emph{CVPR}, 2020.
    
    \bibitem{neuralastar2021}
    R.~Yonetani, T.~Taniai, M.~Barekatain, M.~Nishimura, and A.~Kanezaki, ``Path
      planning using neural {A}* search,'' in \emph{ICML}, 2021.
    
    \bibitem{gupta2017cognitive}
    S.~Gupta, J.~Davidson, S.~Levine, R.~Sukthankar, and J.~Malik, ``Cognitive
      mapping and planning for visual navigation,'' in \emph{CVPR}, 2017, pp.
      2616--2625.
    
    \bibitem{VINNips2016}
    A.~Tamar, Y.~Wu, G.~Thomas, S.~Levine, and P.~Abbeel, ``{Value Iteration
      Networks},'' in \emph{{NeurIPS}}, 2016.
    
    \bibitem{MoleArxiv2022}
    G.~Bono, L.~Antsfeld, A.~Sadek, G.~Monaci, and C.~Wolf., ``{Learning with a
      Mole: Transferable latent spatial representations for navigation without
      reconstruction},'' \emph{arXiv:2306.03857}, 2023.
    
    \bibitem{kastner2022all}
    L.~Kastner, J.~Cox, T.~Buiyan, and J.~Lambrecht, ``All-in-one: A {DRL}-based
      control switch combining state-of-the-art navigation planners,'' in
      \emph{{ICRA}}, 2022.
    
    \bibitem{schulman2017proximal}
    J.~Schulman, F.~Wolski, P.~Dhariwal, A.~Radford, and O.~Klimov, ``Proximal
      policy optimization algorithms,'' \emph{arXiv preprint}, 2017.
    
    \bibitem{chattopadhyay2021robustnav}
    P.~Chattopadhyay, J.~Hoffman, R.~Mottaghi, and A.~Kembhavi, ``Robustnav:
      Towards benchmarking robustness in embodied navigation,'' \emph{CoRR}, vol.
      2106.04531, 2021.
    
    \bibitem{kadian2020sim2real}
    A.~Kadian, J.~Truong, A.~Gokaslan, A.~Clegg, E.~Wijmans, S.~Lee, M.~Savva,
      S.~Chernova, and D.~Batra, ``Sim2real predictivity: Does evaluation in
      simulation predict real-world performance?'' \emph{IEEE Robotics and
      Automation Letters}, vol.~5, no.~4, pp. 6670--6677, 2020.
    
    \bibitem{he2016deep}
    K.~He, X.~Zhang, S.~Ren, and J.~Sun, ``Deep residual learning for image
      recognition,'' in \emph{CVPR}, 2016.
    
    \bibitem{locobot}
    ``{LoCoBot: An Open Source Low Cost Robot},'' \url{www.locobot.org}.
    
    \bibitem{DBLP:conf/nips/WaniPJCS20}
    S.~Wani, S.~Patel, U.~Jain, A.~X. Chang, and M.~Savva, ``Multi{ON}:
      Benchmarking semantic map memory using multi-object navigation,'' in
      \emph{NeurIPS}, 2020.
    
    \bibitem{MarzaIROS2022}
    P.~Marza, L.~Matignon, O.~Simonin, and C.~Wolf, ``{Teaching Agents how to Map:
      Spatial Reasoning for Multi-Object Nav.}'' in \emph{{IROS}}, 2022.
    
    \bibitem{DBLP:journals/corr/abs-1807-06757}
    P.~Anderson, A.~X. Chang, D.~S. Chaplot, A.~Dosovitskiy, S.~Gupta, V.~Koltun,
      J.~Kosecka, J.~Malik, R.~Mottaghi, M.~Savva, and A.~R. Zamir, ``On evaluation
      of embodied navigation agents,'' \emph{arXiv}, 2018.
    
    \bibitem{Teichman-RSS-13}
    A.~Teichman, S.~Miller, and S.~Thrun, ``Unsupervised intrinsic calibration of
      depth sensors via {SLAM},'' in \emph{RSS}, 2013.
    
    \end{thebibliography}

\end{document}